\documentclass{article}

\usepackage{arxiv}

\usepackage[square,numbers]{natbib}
\usepackage[utf8]{inputenc} 
\usepackage[T1]{fontenc}    
\usepackage{hyperref}       
\usepackage{url}            
\usepackage{booktabs}       
\usepackage{amsfonts}       
\usepackage{nicefrac}       
\usepackage{microtype}      
\usepackage{graphicx}
\usepackage{amssymb}
\usepackage{amsmath}
\usepackage{booktabs}
\usepackage[ruled,vlined]{algorithm2e}

\usepackage{array}
\newcolumntype{H}{>{\setbox0=\hbox\bgroup}c<{\egroup}@{}}

\DeclareMathOperator{\sign}{sign}

\newcommand{\xin}{x_{in}}
\newcommand{\Xin}{X_{in}}
\newcommand{\xout}{x_{out}}
\newcommand{\Xout}{X_{out}}
\newcommand{\predy}{\hat{y}}

\title{Out-of-Distribution Detection Using Outlier Detection Methods}

\author{
    Jan Diers \\ 
    Friedrich-Schiller-University Jena \\
    Germany \\ 
    \texttt{jan.diers@uni-jena.de}
   \And
    Christian Pigorsch \\ 
    Friedrich-Schiller-University Jena \\
    Germany \\ 
    \texttt{christian.pigorsch@uni-jena.de}
}

\begin{document}
\maketitle

\begin{abstract}
Out-of-distribution detection (OOD) deals with anomalous input to neural networks. In the past, specialized methods have been proposed to reject predictions on anomalous input. Similarly, it was shown that feature extraction models in combination with outlier detection algorithms are well suited to detect anomalous input.  We use outlier detection algorithms to detect anomalous input as reliable as specialized methods from the field of OOD. No neural network adaptation is required; detection is based on the model's softmax score. Our approach works unsupervised using an Isolation Forest and can be further improved by using a supervised learning method such as Gradient Boosting.
\end{abstract}


\section{Introduction}

Machine learning methods and in particular neural networks are the backbone of many modern applications in research and industry. To apply these methods, at first, training data is collected, then models are trained and evaluated on validation or test data. If the error is low and the model makes reliable predictions, the model is released to be used in software. In the software, for example, it controls autonomous systems, detects production errors, or analyzes text.

But what happens if the data from the real environment does not match the training data well? What happens if the real data does not match the training data, e.g.~if targets have changed? The model will not refuse to predict, but will fail the task undetected. It cannot signal the user that the new input is an unknown class or from a different distribution of the class.

Most of the modern models have no way to detect this change of environment (called "domain adaptation" or "concept drift"). Depending on how the distribution of the test data changes, it is possible to reweight the training data in the model's objective function to reflect the domain change \cite{KunZhang.2013}. However, this does not work if the data and labels change arbitrarily, i. e., no knowledge about the distribution of the test data exists in advance. In practice, unfortunately, this is the case because it is impossible to know in advance what data points the model will be queried on. It is an open challenge for AI safety to teach models not to make a decision if they are unsure about the decision \cite{Amodei.2016}. Current models lack this capability.

Recently, the research field of out-of-distribution detection (OOD) has emerged, which deals with processing data points that do not match the distribution of the training data. The goal is to detect anomalous input and deny an invalid prediction. The raw softmax score is not suitable in these cases for neural networks \cite{ChuanGuo.2017b} as well as for other machine learning methods \cite{NiculescuMizil.2005} - although it is supposed to represent the confidence of the decision.

For image classification, the problem is formulated as follows. A model was trained to distinguish between $M$ different classes $K_1, K_2, ..., K_M$. At inference time, data appears which originates from a different domain, i.e.~not from the $M$ classes already known before, but from a new class. The model can therefore only misclassify, because the true label $K_{M+1}$ is unknown to the model. So how do we need to change models to let them handle anomalous input? How can we find out if an input is from an unknown class?

Hendrycks and Gimpel \cite{Hendrycks.2016} provide a baseline against which misclassified and out-of-domain input can be detected. They empirically find that the predicted confidence of a neural network is lower when the input contains a foreign class $K_{M+1}$. 

It has been shown that extracted features of a neural network trained on ImageNet, provide an effective way to detect anomalous images in the input \cite{Bergman.2020}. In addition, there is evidence that anomalous input in image data can also be reliably detected with supervised learning methods \cite{Ruff.30.05.2020, Hendrycks.2018}. The label for the supervised methods is generated automatically based on arbitrary images that represent the distributions different to the normal distribution.

We follow up this work and show that outlier detection algorithms also reliably detect anomalous input when the neural network has been specifically trained for a particular task. The detection of out-of-distribution data becomes even more reliable when a supervised learning method is used instead of the unsupervised outlier-detection algorithm.

Our proposed methodology does not require re-training of models and can be applied to any existing models. We build on the work of \cite{Hendrycks.2016} and detect OOD input based solely on predicted class probabilities. To do this, we fit an Isolation Forest that separates expected class probabilities from abnormal class probabilities. This enables the detection of the OOD input. Validation data, which should be available for every model anyway, is sufficient for fitting the Isolation Forest.

Our contributions to the research area are:
\begin{enumerate}
    \item We propose an unsupervised method to distinguish in-distribution from out-of-distribution input. The results indicate that the assumptions and methods of outlier and deep anomaly detection are also relevant to the field of out-of-distribution detection.
    \item The method works on the basis of an Isolation Forest. It can be applied to existing models, requires no special training and no special architecture of the model. The employed loss function also does not need to be adapted. Classification accuracy does not suffer since no change is made to the model.
    \item We show empirically, using common benchmark datasets, that our approach leads to a better detection of out-of-domain input than with current OOD detection techniques. This is especially the case when the classification accuracy of the network is already low.
    \item We present that the results can be further improved if a supervised learning method is used instead of the Isolation Forest to detect the anomalies. The supervised learning method generalizes well to previously unknown OOD data.
\end{enumerate}

\section{Related Literature}

The related literature for this work is primarily guided by two research areas. The first is the field of out-of-distribution detection. The second research field covers the area of outlier detection and has received little attention in the field of out-of-distribution detection. We will give a brief overview of the two research fields in the following section. A comprehensive overview is provided by  Saikiran Bulusu et al. \cite{SaikiranBulusu.2020} in their survey of the field of OOD detection.

In the following, we assume that a model $f$ has been trained to predict the probability $f(\xin) = p(y|\xin)$. The estimation of the likelihood will succeed if $x \in \Xin$, i.e., the input $x$ corresponds to the distribution of the training data $\Xin$.

Out-of-distribution detection considers the case where $p(y|\xout)$ is estimated, where $\xout$ is taken from a distribution that has no correspondence with $\Xin$. In particular, the class probabilities $p(y|\xout)$ must be false, since none of the $y$ in $\Xout$ exists. 

Consider the example, that $f$ was trained to distinguish cats from dogs. $\Xin$ then includes images of cats and dogs. It follows that $\Xout$ represents all other images, that neither contain cats nor dogs. The target $y \in \{0, 1\}$ indicates whether the image contains a cat or a dog. The predicted probability $p(y|\xout)$ is then false in all cases, since $\Xout$ does not include images of cats or dogs.

An approach to determine whether $p(y| \cdot )$ was estimated based on $\xin$ or $\xout$ is provided by \cite{Hendrycks.2016}. The paper finds that the overall estimated confidence for $\Xout$ is lower than the estimated confidence for $\Xin$. This allows to define a threshold that detects out-of-domain input: If $p(y| \cdot )$ is too low, then assume $\xout$, otherwise it is $\xin$. Liang et al. \cite{Liang.2017} follow up on this work and propose their method called ODIN. ODIN uses temperature scaling to obtain a better calibrated model that estimates $p(y|\cdot)$ more reliably. This further increases the difference between predictions on $\Xin$ and $\Xout$. In addition, the approach also makes changes to the input to obtain a more robust estimate. To achieve this, the gradient w.r.t.~input is calculated and the input is changed so that $p(y|x)$ increases:
\begin{equation}
    \label{attack}
    \tilde{x} = x - \epsilon \cdot  \sign(-\nabla_x \log(f(x)_y)).
\end{equation}

These studies rely on the difference of the confidences to be sufficiently large for detecting OOD input. In addition, there are other approaches that do not rely on the predicted confidences, such as the one proposed by  DeVries and Taylor \cite{DeVries.2018}. They add an additional output to the model to represent the confidence of the decision. This gives the model the ability to output a confidence for which it expects the estimated probabilities to be correct.

Ren et al. \cite{Ren.07.06.2019} work with two different models. For this purpose, they decompose $p(x)$ into a semantic part $p(x_S)$, which contains the relevant information for the class membership $y$. The irrelevant part (noise), which is not necessary for estimating $p(y|x)$, is subsumed under $p(x_B)$. The joint occurrence is then modeled as $p(x)=p(x_S)\cdot p(x_B)$. The decomposition is used to estimate two different models. The models allow to separate between semantics and noise, which enables the detection of OOD input.

Another line of research has emerged around the name Outlier Exposure. This refers to methods that have already been trained on data from $\Xout$. It turns out that neural networks usually generalize to other, previously unknown data from $\Xout$ and thus provide reliable detection.

Hendrycks et al. \cite{Hendrycks.2018} use Outlier Exposure to adjust the loss function $L$ of the model so that the entropy $p(y|\xout)$ is high. To achieve this, they introduce an additional term into the loss function for the classification task:
\begin{equation*}
\begin{split}
\min 
\mathbb{E}_{(x,y) \sim \Xin}[ L(f(x), y)] 
+ \lambda \mathbb{E}_{x \sim \Xout} [H(f(x), U)]
\end{split},
\end{equation*}
where $H$ corresponds to the cross entropy and $U$ corresponds to the uniform distribution over $M$ classes. The change of the loss function encourages the model to predict a uniform distribution when there is input from $\xout$. The authors also present ways to apply the method when the problem is not a classification task.

Previous work uses datasets $\Xout$, which are semantically different from $\Xin$, to evaluate their methods. Chen et al. \cite{Chen.21.03.2020} note that this evaluation is incomplete. They show that most methods inadequately recognize when OOD is input generated based on adversarial attacks. The authors therefore extend the approach of \cite{Hendrycks.2018} to include data that was generated using adversarial attacks. This produces more robust detection of $\Xout$ when $\Xout$ is minimally modified data from $\Xin$. They use the following adjusted loss function:

\begin{equation*}
\begin{split} 
\min 
\mathbb{E}_{(x,y) \sim \Xin} [\max_{\delta \in B(x,\epsilon)}[-\log(f(x+\delta)_y)]] 
+ \lambda \mathbb{E}_{x \sim \Xout} \max_{\delta \in B(x,\epsilon)}[H(f(x), U)]] .
\end{split}
\end{equation*}

The set of changes to the input that deviate at most by $\epsilon$ from the original input is denoted by $B(x, \epsilon) = \{ \delta \in \mathbb{R}^n:~|\delta||_\infty \leq \epsilon \land x + \delta \text{ is valid}\}$. $x+\delta$ is valid if the minimum and maximum pixel values of the image are maintained. The generation is thus similar to $\tilde{x}$ from \eqref{attack}, with the change in maximum distance from the original $\epsilon$.

Our work uses outlier detection methods to detect the OOD input of neural networks. A fixed definition of outliers does not exist, however, the community has widely agreed on the definition of Hawkins \cite{.hawkins}: \textit{"An outlier is an observation which deviates so much from the other observations as to arouse suspicions that it was generated by a different mechanism"}. The distinction between $\Xin$ and $\Xout$ thus fulfills all requirements to be an application of outlier detection.

Outlier detection methods search for points that are located in areas of low density. If the data generating distribution is known, it is easy to calculate $p(x)$. For empirical data, methods must be found to estimate $p(x)$.

Common methods for outlier detection are based on kernel density estimates \cite{JooSeukKim.2008, LonginJanLatecki.2007, Schubert.2014}, k-nearest neighbor methods (kNN) \cite{FabrizioAngiulli.2002, Schubert.2014}, dimension reduction \cite{Xu.2010, Hoffmann.2007} or support vector machines \cite{.1999, Tax.2004}, among others.  Methods from the domain of neural networks rely, among others, on generative adversarial networks \cite{Schlegl.17.03.2017, Zenati.17.02.2018} or autoencoders \cite{Chen.2017, Zhou.2017} for this purpose. The usage of k-nearest neighbors is an indirect estimation of the local density, since large distances to neighbors correspond to low density in the area.

Successful concepts from other areas of machine learning are also transferred to outlier detection. For example, there are approaches for ensemble learning \cite{Chen.2017b, Zimek.2014, Zimek.2014b, Zimek.2013}, active learning in the case of partially labeled data \cite{Abe.2006, Liu.2019, Trittenbach.2019, Trittenbach.2021} or self-supervised learning to form a supervised task from the unlabeled data \cite{Diers.2021, Li.2021, Sehwag.2021}.

In our work, we use the Isolation Forest \cite{Liu.2008} as a method for outlier detection. As a tree-based method, it has similarity to the implicit density estimation of kNN-based methods with the difference that it is a model-driven approach. The Isolation Forest scales linearly with the size of the data set, which is a major advantage over instance-based kNN methods.

The Isolation Forest is composed of multiple Isolation Trees that perform random splits to features. The objective is to isolate individual data points in the nodes. The fewer splits are required before a point can be isolated into a terminal node, the larger is the outlier factor of that point. The outlier factor is defined as the expected path length of a point in all trees of the forest.

Combining outlier detection methods with neural network features has already been used in the community. For example, \cite{Bergman.2020} use the features of a network trained on ImageNet to detect anomalies using kNN. Our approach is similar. However, unlike \cite{Bergman.2020}, we do not use features based on ImageNet but train each of the networks for the dataset. Furthermore, we do not use intermediate features, but rather the output of the network. Both of these are motivated by the fact that we are interested in detecting OOD input when the network has been trained for a specific task.

The traditional methods of outlier detection work unsupervised. The reason is that outliers from the past are not necessarily representative for outliers in the future. This prohibits the reliable use of supervised learning methods. However, it is not true for every anomaly detection task that the distribution of outliers may change over time. There are some anomalies, e.g. in medicine, whose origin is well understood and therefore it can be assumed with a high degree of certainty that anomalies will not change in the future.

If this is the case, then anomaly detection can be transformed into a supervised problem. The task then corresponds to an ordinary classification problem. Usually, the data is highly unbalanced, since the outliers represent, by definition, a minority of the data points.

For OOD detection, there is evidence that anomalies also do not change over time. For example, Ruff et al. \cite{Ruff.30.05.2020} and Hendrycks et al. \cite{Hendrycks.2018} report that neural networks also detect previously unknown OOD data well when trained with arbitrary, other OOD data. This also justifies the use of outlier exposure models for the task and suggests that supervised learning methods may be well-suited for OOD detection. In our experiments, we use Gradient Boosting \cite{Friedman.2002} for this purpose and confirm that reliable OOD detection is possible with supervised learning methods.

\section{Method}

One of the disadvantages of popular outlier detection methods is that they are designed for tabular data. For this reason, the algorithms cannot be applied to raw image data, but the data must be transformed properly.

For the transformation, we use the mapping of the function $f$ to the output $\predy$. $\predy$ represents the softmax values of a neural network $f$. The values in $\predy$ follow a distribution, which we denote as $ D_{in} $, so that $f(\Xin) \sim D_{in}$. A key assumption in our approach is that the softmax values of the neural network are informative with respect to the question whether a given input value is out of distribution or not. This is a reasonable presumption since a well trained neural network will assign elements from the trained distribution, i.e.~$f(\Xin) \sim D_{in}$, to a given class whereas elements out of distribution, i.e.~$f(\Xout) \sim D_{out}$, will result in values that do not allow for a clear decision, e.g.~the softmax values are very close together. We will come back to this topic in the discussion of our empirical results in Chapter 6.

We use different approaches to distinguish elements of $D_{in}$ from elements of $D_{out}$. Our first approach is based on an unsupervised Isolation Forest. For this, we use the Isolation Forest to estimate the normal distribution $D_{in}$, which we can use to estimate $p(f(x))$. If the Isolation Forest signals that $p(f(x))$ is low, we assume that $x$ does not come from $\Xin$ and instead represents OOD input.

We can determine the distribution $D_{in}$ by splitting a validation set from the training data. On the validation set, we can apply $f$ to obtain data that follow the distribution $D_{in}$. We then fit the Isolation Forest on this data. See also Algorithm \ref{algoSingleForest}.

\begin{algorithm}
\SetAlgoLined
\KwResult{Isolation Forest for OOD-Detection}
\KwIn{Trained neural network, validation set, Isolation Forest}
 
 \Begin{
    1. make predictions on validation set\;
    2. fit Isolation Forest on predictions\;
    3. apply Isolation Forest to new predictions\;
 }
 \caption{OOD-Detection using an Isolation Forest}
 \label{algoSingleForest}
\end{algorithm}

As a second approach, we propose to use supervised learning to distinguish $ D_{in} $ and $ D_{out} $. To do this, we also use the validation data from the training set and take completely different data to simulate $ D_{out} $. In our case, the Food101 dataset \cite{LukasBossard.2014} is the basis for determining $ D_{out} $. The Food101 dataset contains images from different food categories and has no similarity to the other datasets we use.\footnote{Previously published work in the area of OOD often uses the Tiny Images dataset \cite{Torralba.2008} to calibrate the models. This dataset has been withdrawn by the authors \cite{withdrawTinyImages}. Of course, we follow the decision and no longer use this data in our work either. Instead, we use the Food101 \cite{LukasBossard.2014} dataset to train outlier exposure models and the Gradient Boosting classifier.} This procedure is identical to the outlier exposure models. We then fit a Gradient Boosting classifier that separates between points $ f(\Xin) $ and $ f(\Xout) $. The choice of Gradient Boosting is not of particular importance; many classification models are suitable for this purpose. The use of supervised anomaly detection seems feasible because known OOD input generalizes well to unknown OOD input \cite{Ruff.30.05.2020, Hendrycks.2018}. That is, although we use Food101 as an exemplary OOD dataset to train the model, we can also detect OOD input from other datasets. The details are shown by Algorithm \ref{supervisedOOD}.

\begin{algorithm}
\SetAlgoLined
\KwResult{Gradient Boosting Classifier for OOD-Detection}
\KwIn{Trained neural network,\newline validation set,\newline any OOD-dataset (e.g. Food101),\newline Gradient Boosting Classifier (GBM)}
 
 \Begin{
    1. $\predy_{in} \leftarrow $ make predictions on validation set\;
    2. $\predy_{out} \leftarrow $ make predictions on OOD set\;
    3. fit the GBM to learn difference between $\predy_{in}$ and $\predy_{out}$\;
    4. apply GBM to new predictions\;
 }
 \caption{OOD-Detection with a Gradient Boosting Classifier}
 \label{supervisedOOD}
\end{algorithm}

\section{Experiment Setup}

For our experiments, we use 8 different datasets as in- and OOD data. The images are all cropped to a size of $224 \times 224$ pixels and enriched with smaller augmentation steps (flip and contrast). The details of the datasets can be found in Table \ref{detailsDataset}. We use EfficientNets \cite{Tan.} pre-trained on ImageNet data to perform the transfer learning on the respective dataset. We set the label smoothing parameter \cite{Szegedy.2016} to $0.2$. We first train the front layers for three epochs with a learning rate of $0.01$, before further fine-tuning the model starting from the second block with a learning rate of $0.001$. Adam \cite{Kingma.22.12.2014} is used as optimizer. We apply early stopping and reduce the learning rate by a factor of $0.6$ if the validation loss could not be reduced $3$ epochs in a row. The classification error of the models on the different datasets is shown in Table \ref{aggCLFResult}. Note that the methods Baseline, Isolation Forest and Gradient Boosting Classifier provide the same results by design.

\begin{table*}
\centering
\caption{Details for the datasets used in this study. Textures and SVHNCropped are only used as out-of-distribution datasets and therefore do not have a training or validation split. The Food101 dataset is only used as training and validation set to train Outlier Exposure and Gradient Boosting models. It is not used for out of distribution detection on test images.}
\label{detailsDataset}
\begin{tabular}{lrlll}
\toprule
{} & \multicolumn{4}{c}{number of} \\
{} &   classes & train samples & val samples & test samples \\
name        &           &               &             &              \\
\midrule
Cars196     &       196 &          6.515 &        1.629 &         8.041 \\
Cassava     &         5 &          5.656 &        1.889 &         1.885 \\
CatsVsDogs  &         2 &         13.957 &        4.653 &         4.652 \\
Cifar10     &        10 &         40.000 &       10.000 &        10.000 \\
Cifar100    &       100 &         40.000 &       10.000 &        10.000 \\
Food101     &       101 &         60.600 &       15.150 &            - \\
Textures    &        47 &             - &           - &         5.640 \\
SVHNCropped &        10 &             - &           - &        10.000 \\
\bottomrule
\end{tabular}
\end{table*}

In the benchmark we compare 5 different methods: the Baseline approach \cite{Hendrycks.2016b}, ODIN \cite{Liang.2017}, Outlier Exposure \cite{Hendrycks.2018} and our proposed methods based on Isolation Forest and Gradient Boosting.

\begin{table}
\centering
\caption{Classification error on datasets for various methods.}
\label{aggCLFResult}
\begin{tabular}{llr}
\toprule
        &                              &  classification error \\
dataset & method &                       \\
\midrule
CatsVsDogs & Baseline &                  0.01 \\
        & ODIN &                  0.01 \\
        & Outlier Exposure &                  0.01 \\
        & Isolation Forest &                  0.01 \\
        & Gradient Boosting Classifier &                  0.01 \\
Cifar10 & Baseline &                  0.03 \\
        & ODIN &                  0.03 \\
        & Outlier Exposure &                  0.03 \\
        & Isolation Forest &                  0.03 \\
        & Gradient Boosting Classifier &                  0.03 \\
Cifar100 & Baseline &                  0.15 \\
        & ODIN &                  0.15 \\
        & Outlier Exposure &                  0.15 \\
        & Isolation Forest &                  0.15 \\
        & Gradient Boosting Classifier &                  0.15 \\
Cars196 & Baseline &                  0.34 \\
        & ODIN &                  0.34 \\
        & Outlier Exposure &                  0.33 \\
        & Isolation Forest &                  0.34 \\
        & Gradient Boosting Classifier &                  0.34 \\
Cassava & Baseline &                  0.14 \\
        & ODIN &                  0.14 \\
        & Outlier Exposure &                  0.14 \\
        & Isolation Forest &                  0.14 \\
        & Gradient Boosting Classifier &                  0.14 \\
\bottomrule
\end{tabular}
\end{table}

The neural networks are implemented in Tensorflow based on the publications of \cite{Chen.21.03.2020} and \cite{Liang.2017}. For the Isolation Forest and Gradient Boosting, we use the implementations and hyperparameters given by scikit-learn \cite{FabianPedregosa.2011} in version 0.24. We do not perform any further optimization of the hyperparameters. Our code is publicly available on GitHub.\footnote{https://github.com/jandiers/ood-detection}

\section{Results}

For the metrics, we stay consistent with other work in the field. We consider the out-of-distribution error (OOD error), the area-under-curve value (AUC), and the false precision rate when the true precision rate is $95\%$ (FPR at 95\% TPR). We would like to point out that OOD detection is a topic with a high level of practical relevance. The system must make a \textit{binary} decision whether the input is anomalous or not - the score of a point is not sufficient for this task. Therefore, we emphasize the importance of the OOD error, which is the number of misclassified OOD images. In practice, we would like to minimize this error. The AUC is a metric that does not require a threshold but works based on scores.

The results with an Isolation Forest provide comparable results to the methods designed specifically for out-of-distribution detection. The advantage of Isolation Forests is that no adjustment to the model is necessary. If an existing model is already in use, only a validation set is needed to apply our method. ODIN can also be applied to existing models, but this requires two runs: In the first run, the gradient to the input data must be computed, and then in the second run, the confidence on the modified images must be obtained. This significantly increases the runtime of the inference and thus complicates the practical use of the method. The Baseline approach can also be used without modification, but falls behind the results of Isolation Forest and Gradient Boosting. Outlier exposure approaches cannot be applied to existing models because special loss functions must be used during training.

\begin{table}
\centering
\caption{Results averaged over all out-of-distribution datasets and all in-distribution datasets. The supervised OOD-method based on Gradient Boosting classification outperforms all other methods in all metrics. Especially in terms of OOD-error, which is the most important metric when OOD is applied in practice, the supervised methods is clearly the best. Also the OOD detection based on Isolation Forests works well. Note that Isolation Forests work completely unsupervised and therefore solve a much more difficult task to detect OOD-Input.}
\label{allAvgOODResult}
\begin{tabular}{lrrr}
\toprule
{} & \multicolumn{3}{l}{average over 6 OOD datasets} \\
{} &                   OOD error & OOD AUC & FPR at 95\% TPR \\
Average over 5 in-distribution datasets &                             &         &                \\
\midrule
Baseline                                &                        0.30 &    0.85 &           0.51 \\
ODIN                                    &                        0.25 &    0.82 &           0.44 \\
Outlier Exposure                        &                        0.30 &    0.86 &           0.50 \\
Isolation Forest                        &                        0.26 &    0.85 &           0.56 \\
Gradient Boosting Classifier            &                        0.14 &    0.92 &           0.25 \\
\bottomrule
\end{tabular}
\end{table}

The supervised method based on Gradient Boosting works particularly reliably. In all metrics this approach performs best, especially with respect to the OOD error, the superiority of the approach is clear. On average 25\% misclassified OOD input from ODIN can be reduced by Gradient Boosting to an error rate of 14\%, which is an improvement of 44\%.

The AUC tests different thresholds for classification. Most models yield similar AUC values with different OOD error values. This suggests that the threshold for binary classification (OOD vs. in-distribution) is difficult to choose for many methods. Gradient Boosting is again the superior methodology, but by a smaller margin compared to the other metrics.

The third metric of FPR at 95\% TPR is also borrowed from common literature in this research field. It measures the False Positive Rate when the True Positive Rate is 95\%.  A lower value is better here. In this metric, other methods perform slightly better than Isolation Forest. Again, the best model by far is the supervised OOD detection via Gradient Boosting Classifier.

Table \ref{aggOODResult} summarizes the results. Shown are the metrics, each as an average over 6 out-of-distribution datasets. For example, the Isolation Forest has an OOD error of 18\% on the Cifar10 dataset. This means on average 18\% of the images were misidentified as Cifar10 images, even though they were from a different dataset. The baseline approach also has an error rate of 18\% on Cifar10. The error rate for ODIN is 21\%, outlier exposure has 15\%, Gradient Boosting has 20\%.

We list the detailed results for each dataset in the appendix of this paper.

\begin{table}
\centering
\caption{Results for out of distribution detection based on 6 different datasets. Values represent averages. While ODIN and Outlier Exposure require explicit optimization for out of distribution detection, our approach only requires a validation set to learn the outlier distribution. It works with any existing classifier without modification.}
\label{aggOODResult}
\begin{tabular}{llrrr}
\toprule
in-distribution        &   method                           & \multicolumn{3}{l}{average over 6 OOD datasets} \\
dataset        &                              &                   OOD error & OOD AUC & FPR at 95\% TPR \\
\midrule
CatsVsDogs & Baseline &                        0.04 &    0.98 &           0.02 \\
        & ODIN &                        0.05 &    0.98 &           0.04 \\
        & Outlier Exposure &                        0.03 &    0.99 &           0.02 \\
        & Isolation Forest &                        0.09 &    0.98 &           0.03 \\
        & Gradient Boosting Classifier &                        0.04 &    0.98 &           0.07 \\
Cifar10 & Baseline &                        0.18 &    0.89 &           0.37 \\
        & ODIN &                        0.21 &    0.81 &           0.42 \\
        & Outlier Exposure &                        0.15 &    0.93 &           0.29 \\
        & Isolation Forest &                        0.18 &    0.92 &           0.30 \\
        & Gradient Boosting Classifier &                        0.20 &    0.91 &           0.29 \\
Cifar100 & Baseline &                        0.25 &    0.86 &           0.50 \\
        & ODIN &                        0.21 &    0.86 &           0.40 \\
        & Outlier Exposure &                        0.23 &    0.88 &           0.47 \\
        & Isolation Forest &                        0.38 &    0.71 &           0.94 \\
        & Gradient Boosting Classifier &                        0.27 &    0.87 &           0.43 \\
Cars196 & Baseline &                        0.45 &    0.66 &           0.96 \\
        & ODIN &                        0.43 &    0.61 &           0.93 \\
        & Outlier Exposure &                        0.45 &    0.68 &           0.95 \\
        & Isolation Forest &                        0.44 &    0.78 &           0.75 \\
        & Gradient Boosting Classifier &                        0.05 &    0.99 &           0.04 \\
Cassava & Baseline &                        0.57 &    0.85 &           0.69 \\
        & ODIN &                        0.35 &    0.86 &           0.43 \\
        & Outlier Exposure &                        0.62 &    0.80 &           0.76 \\
        & Isolation Forest &                        0.22 &    0.86 &           0.77 \\
        & Gradient Boosting Classifier &                        0.12 &    0.88 &           0.40 \\
\bottomrule
\end{tabular}
\end{table}

\section{Discussion}

An open question is why the anomalous input can be detected in the softmax values of the neural network. Although it is consistent with previous research that the softmax values allow good generalization to other OOD input \cite{Hendrycks.2018, Ruff.30.05.2020}, there is a lack of arguments to support these observations. In the introduction of our method, we stated that $f(\Xin) \sim D_{in}$ and consequently $f(\Xout) \sim D_{out}$ holds and that the corresponding softmax values allow for a distinction between the elements of these distributions. To support this assumption and to visualize the distributions we compute the t-SNE visualization of the softmax activations of in-distribution data ($D_{in}$) and plot the softmax activations of out-of-distribution data ($D_{out}$) against it.

\begin{figure*}
\label{outputDist}
\centering
\includegraphics[width=\textwidth]{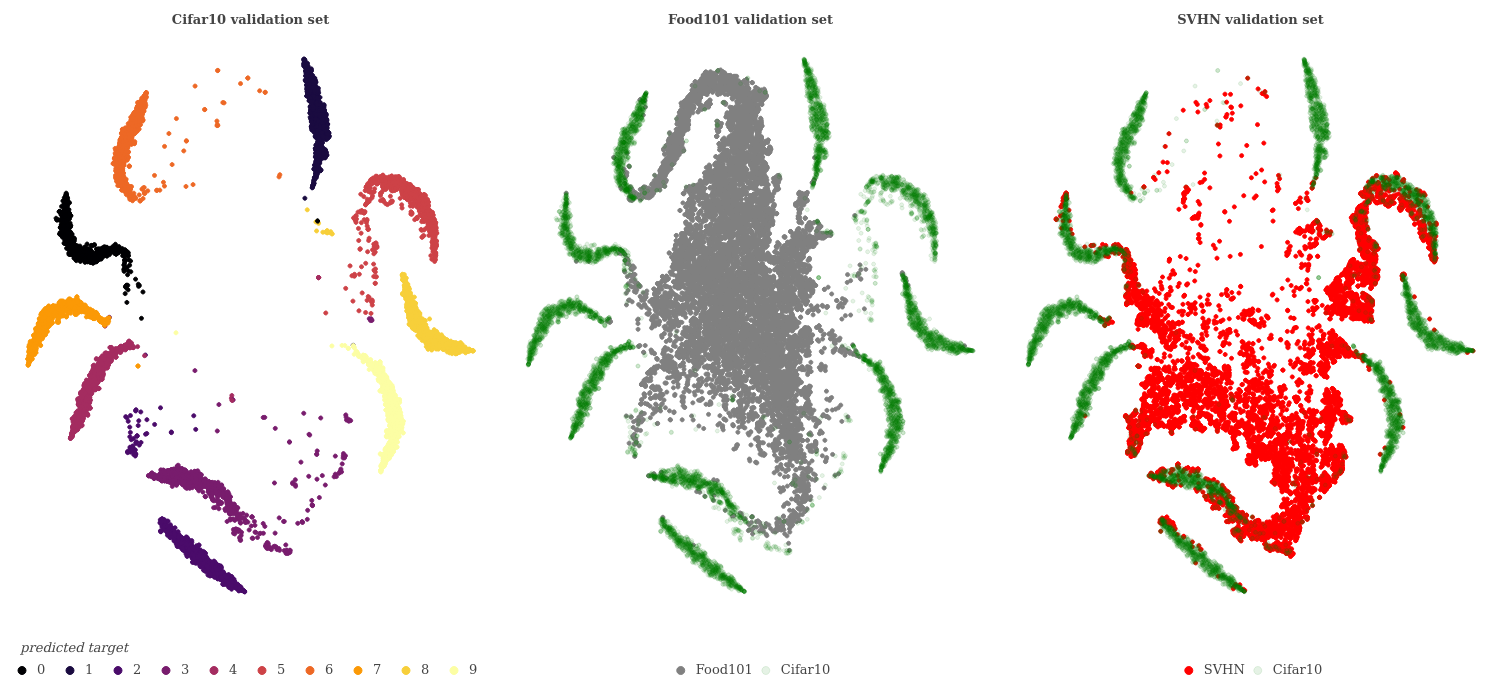}
\caption{The softmax activations of the neural network reduced to 2 dimensions. As expected, on the left, the predicted classes form clearly separated clusters. Between these clusters lie the Food101 and SVHN OOD datasets, respectively. SVHN is projected at a similar location as Food101, which explains the good results of the supervised outlier detection. This does not mean that points are also projected onto each other in the full dimension, however there is a clear separation from the in-distribution data. Gradient Boosting learns to distinguish between Food101 and Cifar10, but can also distinguish SVHN from Cifar10 with it.}
\end{figure*}

Taking Cifar10 as $\Xin$ and Food101 or SVHN as $\Xout$ as examples, we can see in Figure \ref{outputDist} that the predicted classes of Cifar10 form well-formed clusters (leftmost figure). Between these clusters is a wide set of points that do not belong to any cluster. These are the predictions on Food101 (center figure) and SVHN (right figure), respectively. It can be observed that even previously unknown OOD data is projected where already known OOD data is located. While t-SNE does not reflect a global topology, it is a clear indication that the clusters of in-distribution and OOD data are distinctly separated. This explains the absence of the distribution shift in the neural network output and enables supervised detection of OOD input. The performance of the Isolation Forest is explained by the well-defined clusters. The cluster structure represents a fundamental assumption of outlier detection that is exploited by the Isolation Forest. In existing clusters (in-distribution clusters), it is difficult for the Isolation Forest to isolate individual points. In widely spreaded clusters, this is easier (out-of-distribution data).

When looking at the results, it is also noticeable that Gradient Boosting is superior to the other methods. One reason might be, that datasets that are difficult to classify (Cassava, Cars196) do not form a good precondition for existing OOD methods. If general classification accuracy already suffers, then OOD detection is particularly difficult. It also turns out that using a model on the entire softmax distribution is beneficial. The input to the Baseline approach is exactly the same as for Gradient Boosting and Isolation Forest, however these methods are superior to the simple Baseline.

For future research it is of interest to combine performant approaches for OOD detection. ODIN or Outlier Exposure can be used as a basis to apply Isolation Forests or Gradient Boosting on the softmax values. It is expected that this will further improve performance. Furthermore, it seems challenging to include datasets in the benchmark comparisons that contain real-world OOD data. In OOD detection, autonomously acting systems are the focus of interest. Thus, for future research, we plan to use images of autonomously acting agents. By doing so, we would like to test whether OOD methods can contribute to agents' ability to better orient and act in previously unknown environments.

\bibliography{citation}  

\newpage
\section{Appendix}
\setcounter{table}{0}
\renewcommand{\thetable}{A\arabic{table}}

\begin{table}[h]
    \centering
    \caption{Detailed results for the OOD-detection using CIFAR10 as in-distribution data. The OOD threshold defines the minimum confidence that is required to assign a data point to the in-distribution. If the confidence is lower than the score, then the data point is assigned to the OOD distribution. The score is calculated as the 0.05 quantile of the in-distribution scores, which means that 95\% of all in-distribution scores are greater than the threshold.}
\begin{tabular}{lllHccc}
\toprule
            OOD data & method                              & OOD threshold & classification error & OOD error & OOD AUC & FPR at 95\% TPR \\
\toprule
CatsVsDogs & Baseline &        0.7381 &               0.0267 &    0.2787 &   0.829 &         0.7704 \\
            & ODIN &        0.4144 &               0.0303 &    0.2471 &  0.6982 &         0.6707 \\
            & Outlier Exposure &        0.7281 &                0.027 &    0.2178 &  0.8602 &         0.5785 \\
            & Isolation Forest &             - &               0.0267 &     0.323 &  0.7524 &         0.8661 \\
            & Gradient Boosting Classifier &             - &               0.0267 &    0.3151 &   0.817 &         0.5132 \\
\midrule
Cifar100 & Baseline &        0.7381 &               0.0267 &    0.1798 &  0.8882 &         0.3096 \\
            & ODIN &        0.4144 &               0.0303 &     0.278 &  0.7779 &         0.5061 \\
            & Outlier Exposure &        0.7281 &                0.027 &    0.1687 &  0.9027 &         0.2871 \\
            & Isolation Forest &             - &               0.0267 &      0.23 &  0.9195 &         0.3169 \\
            & Gradient Boosting Classifier &             - &               0.0267 &    0.2379 &   0.913 &         0.4739 \\
\midrule
Cars196 & Baseline &        0.7381 &               0.0267 &     0.461 &  0.6985 &          0.972 \\
            & ODIN &        0.4144 &               0.0303 &    0.4011 &  0.6787 &         0.8378 \\
            & Outlier Exposure &        0.7281 &                0.027 &    0.3525 &  0.8807 &         0.7283 \\
            & Isolation Forest &             - &               0.0267 &    0.3274 &  0.9421 &            0.4 \\
            & Gradient Boosting Classifier &             - &               0.0267 &     0.448 &  0.7443 &         0.5019 \\
\midrule
Cassava & Baseline &        0.7381 &               0.0267 &    0.0474 &  0.9859 &         0.0334 \\
            & ODIN &        0.4144 &               0.0303 &    0.0477 &  0.9817 &         0.0355 \\
            & Outlier Exposure &        0.7281 &                0.027 &    0.0438 &   0.995 &         0.0106 \\
            & Isolation Forest &             - &               0.0267 &    0.0388 &  0.9889 &         0.0345 \\
            & Gradient Boosting Classifier &             - &               0.0267 &     0.021 &  0.9943 &         0.0162 \\
\midrule
Textures & Baseline &        0.7381 &               0.0267 &    0.0393 &  0.9923 &           0.02 \\
            & ODIN &        0.4144 &               0.0303 &    0.0519 &  0.9687 &         0.0553 \\
            & Outlier Exposure &        0.7281 &                0.027 &    0.0336 &  0.9969 &         0.0044 \\
            & Isolation Forest &             - &               0.0267 &    0.0325 &  0.9944 &          0.017 \\
            & Gradient Boosting Classifier &             - &               0.0267 &     0.022 &  0.9971 &         0.0114 \\
\midrule
SVHNCropped & Baseline &        0.7381 &               0.0267 &    0.0905 &  0.9404 &          0.131 \\
            & ODIN &        0.4144 &               0.0303 &    0.2386 &  0.7738 &         0.4271 \\
            & Outlier Exposure &        0.7281 &                0.027 &    0.0853 &  0.9557 &         0.1205 \\
            & Isolation Forest &             - &               0.0267 &    0.1548 &  0.9484 &         0.1932 \\
            & Gradient Boosting Classifier &             - &               0.0267 &    0.1684 &  0.9638 &          0.201 \\
\bottomrule
\end{tabular}
\end{table}

\begin{table}[h]
    \centering
    \caption{Detailed results for the OOD-detection using CIFAR100 as in-distribution data. For details on how the OOD threshold is defined, see Table A1.}
\begin{tabular}{lllHccc}
\toprule
            OOD data & method                              & OOD threshold & classification error & OOD error & OOD AUC & FPR at 95\% TPR \\
\toprule
CatsVsDogs & Baseline &        0.2129 &               0.1498 &    0.1349 &  0.9392 &         0.3173 \\
            & ODIN &        0.1437 &               0.1498 &    0.1244 &  0.9139 &         0.2844 \\
            & Outlier Exposure &        0.2145 &               0.1485 &    0.1141 &  0.9563 &         0.2519 \\
            & Isolation Forest &             - &               0.1498 &    0.3175 &  0.7291 &         0.9323 \\
            & Gradient Boosting Classifier &             - &               0.1498 &    0.1461 &  0.9625 &         0.1347 \\
\midrule
Cifar10 & Baseline &        0.2129 &               0.1498 &    0.3412 &  0.8389 &         0.6324 \\
            & ODIN &        0.1437 &               0.1498 &    0.3601 &  0.7856 &         0.6703 \\
            & Outlier Exposure &        0.2145 &               0.1485 &     0.347 &  0.8324 &          0.644 \\
            & Isolation Forest &             - &               0.1498 &       0.5 &  0.6351 &         0.9439 \\
            & Gradient Boosting Classifier &             - &               0.1498 &      0.46 &  0.6961 &         0.9485 \\
\midrule
Cars196 & Baseline &        0.2129 &               0.1498 &    0.3771 &  0.8212 &         0.7839 \\
            & ODIN &        0.1437 &               0.1498 &    0.1531 &  0.9417 &         0.2813 \\
            & Outlier Exposure &        0.2145 &               0.1485 &    0.4039 &  0.7975 &          0.844 \\
            & Isolation Forest &             - &               0.1498 &    0.4457 &  0.6698 &         0.9442 \\
            & Gradient Boosting Classifier &             - &               0.1498 &    0.3336 &  0.8917 &         0.4353 \\
\midrule
Cassava & Baseline &        0.2129 &               0.1498 &    0.0708 &  0.9611 &         0.1809 \\
            & ODIN &        0.1437 &               0.1498 &    0.0611 &  0.9687 &         0.1199 \\
            & Outlier Exposure &        0.2145 &               0.1485 &     0.067 &  0.9733 &          0.157 \\
            & Isolation Forest &             - &               0.1498 &    0.1586 &  0.7393 &         0.9406 \\
            & Gradient Boosting Classifier &             - &               0.1498 &    0.0639 &  0.9683 &         0.1331 \\
\midrule
Textures & Baseline &        0.2129 &               0.1498 &    0.1782 &  0.9116 &         0.4055 \\
            & ODIN &        0.1437 &               0.1498 &    0.1402 &  0.9035 &            0.3 \\
            & Outlier Exposure &        0.2145 &               0.1485 &    0.1426 &  0.9353 &         0.3067 \\
            & Isolation Forest &             - &               0.1498 &    0.3606 &  0.7681 &         0.8787 \\
            & Gradient Boosting Classifier &             - &               0.1498 &    0.1659 &  0.9548 &         0.1757 \\
\midrule
SVHNCropped & Baseline &        0.2129 &               0.1498 &    0.3718 &  0.7078 &         0.6936 \\
            & ODIN &        0.1437 &               0.1498 &    0.4001 &  0.6703 &         0.7502 \\
            & Outlier Exposure &        0.2145 &               0.1485 &    0.3216 &  0.8068 &         0.5933 \\
            & Isolation Forest &             - &               0.1498 &       0.5 &  0.6957 &         0.9725 \\
            & Gradient Boosting Classifier &             - &               0.1498 &    0.4721 &  0.7182 &         0.8016 \\
\bottomrule
\end{tabular}
\end{table}

\begin{table}[h]
    \centering
    \caption{Detailed results for the OOD-detection using CatsVsDogs as in-distribution data. For details on how the OOD threshold is defined, see Table A1.}
\begin{tabular}{lllHccc}
\toprule
            OOD data & method                              & OOD threshold & classification error & OOD error & OOD AUC & FPR at 95\% TPR \\
\toprule
Cifar10 & Baseline &         0.862 &               0.0084 &    0.0842 &   0.958 &         0.1004 \\
            & ODIN &        0.7616 &               0.0084 &    0.1181 &  0.9427 &         0.1503 \\
            & Outlier Exposure &        0.8836 &                0.006 &    0.0754 &  0.9615 &         0.0873 \\
            & Isolation Forest &             - &               0.0084 &    0.1007 &  0.9536 &         0.1266 \\
            & Gradient Boosting Classifier &             - &               0.0084 &    0.1193 &  0.9554 &         0.3356 \\
\midrule
Cifar100 & Baseline &         0.862 &               0.0084 &    0.0366 &  0.9839 &         0.0304 \\
            & ODIN &        0.7616 &               0.0084 &    0.0599 &   0.977 &         0.0648 \\
            & Outlier Exposure &        0.8836 &                0.006 &    0.0308 &  0.9882 &         0.0218 \\
            & Isolation Forest &             - &               0.0084 &    0.0718 &  0.9806 &         0.0428 \\
            & Gradient Boosting Classifier &             - &               0.0084 &    0.0581 &  0.9815 &          0.043 \\
\midrule
Cars196 & Baseline &         0.862 &               0.0084 &    0.0187 &  0.9841 &         0.0005 \\
            & ODIN &        0.7616 &               0.0084 &    0.0187 &  0.9972 &         0.0005 \\
            & Outlier Exposure &        0.8836 &                0.006 &    0.0194 &  0.9838 &         0.0016 \\
            & Isolation Forest &             - &               0.0084 &    0.0719 &  0.9812 &         0.0039 \\
            & Gradient Boosting Classifier &             - &               0.0084 &    0.0144 &  0.9903 &         0.0181 \\
\midrule
Cassava & Baseline &         0.862 &               0.0084 &    0.0358 &  0.9879 &         0.0005 \\
            & ODIN &        0.7616 &               0.0084 &    0.0363 &   0.997 &         0.0021 \\
            & Outlier Exposure &        0.8836 &                0.006 &    0.0356 &  0.9904 &            0.0 \\
            & Isolation Forest &             - &               0.0084 &    0.1395 &  0.9858 &          0.009 \\
            & Gradient Boosting Classifier &             - &               0.0084 &    0.0239 &  0.9915 &          0.017 \\
\midrule
Textures & Baseline &         0.862 &               0.0084 &     0.025 &  0.9908 &         0.0043 \\
            & ODIN &        0.7616 &               0.0084 &    0.0271 &  0.9958 &         0.0083 \\
            & Outlier Exposure &        0.8836 &                0.006 &    0.0247 &  0.9918 &         0.0037 \\
            & Isolation Forest &             - &               0.0084 &    0.0897 &  0.9884 &         0.0087 \\
            & Gradient Boosting Classifier &             - &               0.0084 &    0.0204 &  0.9914 &         0.0174 \\
\midrule
SVHNCropped & Baseline &         0.862 &               0.0084 &    0.0159 &  0.9959 &            0.0 \\
            & ODIN &        0.7616 &               0.0084 &    0.0159 &  0.9996 &            0.0 \\
            & Outlier Exposure &        0.8836 &                0.006 &    0.0159 &  0.9978 &            0.0 \\
            & Isolation Forest &             - &               0.0084 &    0.0622 &  0.9925 &            0.0 \\
            & Gradient Boosting Classifier &             - &               0.0084 &    0.0148 &  0.9884 &         0.0155 \\
\bottomrule
\end{tabular}
\end{table}

\begin{table}[h]
    \centering
    \caption{Detailed results for the OOD-detection using Cars196 as in-distribution data. For details on how the OOD threshold is defined, see Table A1.}
\begin{tabular}{lllHccc}
\toprule
            OOD data & method                              & OOD threshold & classification error & OOD error & OOD AUC & FPR at 95\% TPR \\
\toprule
CatsVsDogs & Baseline &        0.0766 &               0.3393 &    0.3611 &  0.6999 &         0.8988 \\
            & ODIN &        0.0593 &               0.3393 &    0.3651 &  0.5945 &         0.9095 \\
            & Outlier Exposure &        0.0683 &               0.3307 &    0.3409 &  0.7895 &         0.8435 \\
            & Isolation Forest &             - &               0.3393 &    0.3665 &  0.7427 &         0.8738 \\
            & Gradient Boosting Classifier &             - &               0.3393 &    0.0295 &  0.9968 &         0.0132 \\
\midrule
Cifar10 & Baseline &        0.0766 &               0.3393 &    0.5657 &  0.6439 &         0.9802 \\
            & ODIN &        0.0593 &               0.3393 &     0.566 &  0.5558 &         0.9808 \\
            & Outlier Exposure &        0.0683 &               0.3307 &    0.5668 &  0.6551 &         0.9822 \\
            & Isolation Forest &             - &               0.3393 &    0.5543 &  0.7833 &         0.7742 \\
            & Gradient Boosting Classifier &             - &               0.3393 &    0.0851 &  0.9819 &         0.0751 \\
\midrule
Cifar100 & Baseline &        0.0766 &               0.3393 &    0.5624 &  0.6514 &         0.9743 \\
            & ODIN &        0.0593 &               0.3393 &    0.5581 &  0.5935 &         0.9665 \\
            & Outlier Exposure &        0.0683 &               0.3307 &     0.568 &  0.6579 &         0.9845 \\
            & Isolation Forest &             - &               0.3393 &    0.5543 &  0.8006 &         0.7523 \\
            & Gradient Boosting Classifier &             - &               0.3393 &    0.0423 &  0.9923 &         0.0358 \\
\midrule
Cassava & Baseline &        0.0766 &               0.3393 &    0.2287 &  0.6356 &         0.9905 \\
            & ODIN &        0.0593 &               0.3393 &    0.2293 &  0.5585 &         0.9936 \\
            & Outlier Exposure &        0.0683 &               0.3307 &    0.2173 &  0.7248 &         0.9305 \\
            & Isolation Forest &             - &               0.3393 &    0.1899 &  0.7277 &         0.8631 \\
            & Gradient Boosting Classifier &             - &               0.3393 &    0.0339 &  0.9949 &          0.023 \\
\midrule
Textures & Baseline &        0.0766 &               0.3393 &    0.4324 &  0.6134 &         0.9773 \\
            & ODIN &        0.0593 &               0.3393 &    0.4082 &  0.5752 &         0.9188 \\
            & Outlier Exposure &        0.0683 &               0.3307 &    0.4294 &  0.6213 &         0.9702 \\
            & Isolation Forest &             - &               0.3393 &    0.4123 &  0.6915 &         0.8674 \\
            & Gradient Boosting Classifier &             - &               0.3393 &    0.1022 &  0.9728 &         0.1024 \\
\midrule
SVHNCropped & Baseline &        0.0766 &               0.3393 &    0.5551 &  0.7013 &         0.9612 \\
            & ODIN &        0.0593 &               0.3393 &    0.4625 &  0.7632 &         0.7941 \\
            & Outlier Exposure &        0.0683 &               0.3307 &    0.5732 &  0.6336 &         0.9938 \\
            & Isolation Forest &             - &               0.3393 &    0.5543 &  0.9219 &         0.3882 \\
            & Gradient Boosting Classifier &             - &               0.3393 &    0.0235 &  0.9968 &         0.0153 \\
\bottomrule
\end{tabular}

\end{table}

\begin{table}[h]
    \centering
    \caption{Detailed results for the OOD-detection using Cassava as in-distribution data. For details on how the OOD threshold is defined, see Table A1.}
\begin{tabular}{lllHccc}
\toprule
            OOD data & method                              & OOD threshold & classification error & OOD error & OOD AUC & FPR at 95\% TPR \\
\toprule
CatsVsDogs & Baseline &         0.408 &               0.1379 &    0.4751 &  0.8502 &         0.6475 \\
            & ODIN &        0.5131 &               0.1379 &    0.4421 &  0.7382 &         0.6012 \\
            & Outlier Exposure &        0.3927 &               0.1448 &    0.5838 &  0.7907 &         0.8003 \\
            & Isolation Forest &             - &               0.1379 &     0.344 &  0.7954 &         0.9643 \\
            & Gradient Boosting Classifier &             - &               0.1379 &    0.1851 &  0.8313 &         0.4806 \\
\midrule
Cifar10 & Baseline &         0.408 &               0.1379 &    0.6475 &  0.8172 &         0.7605 \\
            & ODIN &        0.5131 &               0.1379 &    0.5017 &  0.8057 &         0.5875 \\
            & Outlier Exposure &        0.3927 &               0.1448 &    0.7058 &  0.7817 &         0.8294 \\
            & Isolation Forest &             - &               0.1379 &    0.0919 &  0.9186 &          0.603 \\
            & Gradient Boosting Classifier &             - &               0.1379 &    0.1123 &  0.8764 &         0.4265 \\
\midrule
Cifar100 & Baseline &         0.408 &               0.1379 &    0.6246 &   0.843 &         0.7329 \\
            & ODIN &        0.5131 &               0.1379 &    0.4218 &  0.8563 &         0.4923 \\
            & Outlier Exposure &        0.3927 &               0.1448 &    0.6461 &  0.8023 &         0.7587 \\
            & Isolation Forest &             - &               0.1379 &     0.112 &  0.9071 &         0.6451 \\
            & Gradient Boosting Classifier &             - &               0.1379 &    0.1141 &  0.8701 &         0.4308 \\
\midrule
Cars196 & Baseline &         0.408 &               0.1379 &    0.4654 &  0.8994 &          0.563 \\
            & ODIN &        0.5131 &               0.1379 &    0.2179 &   0.937 &         0.2577 \\
            & Outlier Exposure &        0.3927 &               0.1448 &    0.5629 &  0.8697 &         0.6834 \\
            & Isolation Forest &             - &               0.1379 &    0.3035 &  0.8219 &         0.9799 \\
            & Gradient Boosting Classifier &             - &               0.1379 &    0.0901 &  0.9199 &         0.2557 \\
\midrule
Textures & Baseline &         0.408 &               0.1379 &    0.5552 &  0.8231 &         0.7246 \\
            & ODIN &        0.5131 &               0.1379 &    0.3708 &  0.8267 &         0.4796 \\
            & Outlier Exposure &        0.3927 &               0.1448 &    0.5456 &  0.8142 &         0.7115 \\
            & Isolation Forest &             - &               0.1379 &    0.3935 &   0.775 &         0.9257 \\
            & Gradient Boosting Classifier &             - &               0.1379 &    0.1559 &  0.8642 &         0.4599 \\
\midrule
SVHNCropped & Baseline &         0.408 &               0.1379 &    0.6308 &  0.8789 &         0.7408 \\
            & ODIN &        0.5131 &               0.1379 &    0.1306 &  0.9663 &          0.146 \\
            & Outlier Exposure &        0.3927 &               0.1448 &    0.6561 &  0.7666 &         0.7706 \\
            & Isolation Forest &             - &               0.1379 &    0.0829 &  0.9293 &         0.4846 \\
            & Gradient Boosting Classifier &             - &               0.1379 &    0.0779 &  0.8933 &         0.3358 \\
\bottomrule
\end{tabular}
\end{table}

\end{document}